\def\year{2019}\relax
\documentclass[letterpaper]{article} 
\usepackage{aaai19}  
\usepackage{times}  
\usepackage{helvet}  
\usepackage{courier}  
\usepackage{url}  
\usepackage{graphicx}  
\usepackage{multirow}
\usepackage{amsmath}

\title{3D Object Detection Using Scale Invariant and Feature Reweighting Networks}
\author{Xin Zhao$^1$, Zhe Liu$^{2}$\thanks{Corresponding author}, Ruolan Hu$^2$, Kaiqi Huang$^1$\\
$^1$Center for Research on Intelligent System and Engineering\\
Institute of Automation, Chinese Academy of Sciences, Beijing, China, 100190\\
$^2$Huazhong University of Science and Technology, Wuhan, China, 430074\\
xzhao@nlpr.ia.ac.cn, m201772494@hust.edu.cn, huruolan@hust.edu.cn, kqhuang@nlpr.ia.ac.cn\\
}
\begin{document}
\maketitle
\begin{abstract}
3D object detection plays an important role in a large number of real-world applications. It requires us to estimate the localizations and the orientations of 3D objects in real scenes. In this paper, we present a new network architecture which focuses on utilizing the front view images and frustum point clouds to generate 3D detection results. On the one hand, a PointSIFT module is utilized to improve the performance of 3D segmentation. It can capture the information from different orientations in space and the robustness to different scale shapes. On the other hand, our network obtains the useful features and suppresses the features with less information by a SENet module. This module reweights channel features and estimates the 3D bounding boxes more effectively. Our method is evaluated on both KITTI dataset for outdoor scenes and SUN-RGBD dataset for indoor scenes. The experimental results illustrate that our method achieves better performance than the state-of-the-art methods especially when point clouds are highly sparse.
\end{abstract}

\section{Introduction}
In recent years, remarkable progress has been made on the task of 2D object detection in complex scenes by deep convolutional neural networks~\cite{girshick2015fast,redmon2016you,liu2016ssd}. However, it still remains an open problem on 3D object detection. Such as on the KITTI object detection benchmark~\cite{geiger2012we}, a great gap of the average precision (AP) still remains between 2D and 3D object detections. The most important thing of 3D object detection is the way to use the 3D information which represents the depth information of objects in the estimation task. In addition, different from 2D object detection, we need to consider the orientations of 3D bounding boxes simultaneously. Since 3D Lidar is limited by the number of horizontal scan lines and has a non-uniform sampling manner in 3D space, the obtained point clouds are usually very sparse and uneven distributed. These difficulties on 3D object detection bring us enormous challenges.

Some existing methods~\cite{li2016vehicle,wang2017fusing} project the 3D point clouds to the 2D images or fuse the object information from multiple views~\cite{su2015multi,rubino20183d}. These approaches are intuitive and have obtained satisfactory results in simple scenes. However, they are limited in complex scenes for the reason of losing 3D information. Other methods~\cite{engelcke2017vote3deep,maturana2015voxnet} convert point clouds into a 3D voxel grid by quantization and extract each voxel feature by applying 3D convolutional neural networks. However, these approaches have high computational costs especially on dealing with 3D convolutional operations preventing them for real-world applications.

Since 3D data is usually storaged in the form of point clouds, we can mainly make use of this kind of data to extract shape features. Recently, Qi et al. provide a unified network architecture named Pointnet~\cite{qi2017pointnet}, which can represent the permutation invariance of the original input point clouds. It is widely applied in 3D object classification and 3D part segmentation. Pointnet has strong ability to capture global structure information rather than local information. Therefore, an improved version of Pointnet which called Pointnet++~\cite{qi2017pointnet++} is proposed to obtain local structures by increasing contextual scales in distance metric space.

Although point clouds have abundant 3D information, as the complexity of real scenes, it is still very difficult to localize 3D objects only with point clouds especially when they are highly sparse. It is necessary to combine 3D point clouds with 2D images which can provide comprehensive perception information. Some researchers~\cite{wang2017fusing,ku2018joint} utilize the front view images and 3D point clouds to further improve the accuracy of 3D object detection. On the one hand, 2D front view images have rich appearance information of scenes, but they fail to obtain the spatial descriptions. On the other hand, 3D point clouds have accurate location and reflection intensity information, while 3D point clouds are usually very sparse. Thus, 3D point clouds and 2D images are complementary on 3D object detection tasks. With this observation, F-Pointnet~\cite{qi2017frustum} leverages both effective 2D object detectors~\cite{fu2017dssd,lin2017feature} and advanced 3D deep neural networks (Pointnet/Pointnet++) for 3D object detection and localization. Their method gains the cutting-edge results on both KITTI and SUN-RGBD 3D object detection benchmarks. However, it still has some limitations. Since Pointnet++ uses K-nearest searching method, it may lose many useful local orientation information. In real scenes, objects have quite different scales. Thus, we need to consider the scale variations of 3D objects. In addition, each channel feature has different contribution to the network which should have the ability to distinguish their importance.

In this paper, we propose a sub-network (Point-UNet) to achieve the scale invariance and capture the orientation information. In term of the network structure, we can obtain context information and produce the precise localization through a symmetric expanding path. The network can be trained from limited samples but obtains a satisfactory result. This is also captured in 3D UNet~\cite{cciccek20163d} for segmentation tasks. Since the standard 3D convolutional, pooling and upsampling operations would lead to a large amount of computation cost. In our network, we take the Set Abstraction (SA) module and the point Feature Propagation (FP) module~\cite{qi2017pointnet++} to improve our network. Point-UNet has the ability to learn the different orientation information while it is adaptive to scale variations. We achieve this by integrating a PointSIFT module~\cite{jiang2018pointsift} into our network. Since the front view image has rich appearance information, we encode them into our Point-UNet with Resnet-50~\cite{he2016deep} as our backbone feature extractor. Thus, our network can distinguish the categories of objects efficiently through the color information of each 3D point.
  
The second component of our network is a T-Net sub-network. It is used to centralize the points of interests initially. In the network, we use a SA module not only to learn the global feature but also take the extra Lidar reflection intensity feature into consideration. Finally, Point-SENet is the sub-network designed to estimate final 3D bounding boxes. Since each channel feature has different contribution to the whole network. To make the network have an ability to enhance the useful features and suppresses the useless features, we extend the SENet~\cite{hu2017squeeze} module into 3D space to learn the relationship of 3D point channel features.

For outdoor scenes, F-Pointnet~\cite{qi2017frustum} has obtained 8.04\% Average Precision (AP) better than~\cite{chen2017multi} on 3D detection task for Cars on KITTI dataset. For indoor scenes, it has also achieved 4.4\% better 3D mean Average Precision (mAP) than COG~\cite{ren2016three} on SUN-RGBD dataset. Compared with F-Pointnet~\cite{qi2017frustum}, our model obtains 1.4\% higher 3D mAP performance on the validation set of KITTI and 4.4\% higher 3D mAP performance on the test set of SUN-RGBD.

Our contributions in this paper are as follows:

\begin{itemize}
\item We propose a new network architecture which has scale invariance to the shape of point clouds. It has excellent applicability to highly sparse point clouds as it is robust to different scale shapes.
\item Our network can reweight features by learning the correlation from different channel features. This takes the relationship of different channel features into account.  
\item Our method has significant and consistent improvement on both outdoor dataset and indoor dataset compared with the state-of-the-art approaches.
\end{itemize}

\section{Related Work}
Researchers propose high-quality hand-crafted feature representations to localize 3D objects~\cite{dorai1997cosmos,johnson1999using,rusu2009fast} when 3D object detection just emerges. These methods can obtain acceptable results when 3D shape descriptions are available in a simple scene. However, they fail to learn potential invariant features from  more complex scenes so that they have a certain gap in practical applications.

The monocular RGB images can provide rich appearance and context information. Deep3DBox~\cite{mousavian20173d} finds the fact that the perspective projection of the 3D bounding boxes should fit tightly within their 2D detection boxes. It extracts the 3D bounding boxes only from monocular RGB images. ~\cite{chen2016monocular} proposes an energy minimization method that puts candidate boxes in 3D space by exploiting the fact that objects should be on the ground-plane and perform 3D object detection from monocular images. However, due to lacking of depth information, they fail to get high accuracy on 3D object detection tasks.

Estimating the orientations of 3D bounding boxes is of great importance as the orientation angles will directly affect the 3D detection. We can easily get the orientation information from the bird's eye view of 3D point clouds. ~\cite{chen2017multi} utilizes a bird's eye view representation of point clouds to generate 3D highly accurate candidate boxes. However, this region proposal method would fall behind for small object detection.

3D space can be discretized into a 3D voxel grid and a sliding window method can be used through all three dimensions for detection~\cite{wang2015voting}. It is also demonstrated that the exhaustive window searching in 3D space can be extremely efficient by exploiting the sparsity of the problem. Voxelnet~\cite{zhou2017voxelnet} divides 3D points into 3D voxel grids equally in space, putting forward a novel voxel feature encoding (VFE) layer to encode each voxel via stacked VFE layers, which enable the interaction between points within a voxel by combining point-wise features with a locally aggregated feature. Then, the region proposal network takes the voxel features and obtains the 3D detection results.

In order to reduce the search space, F-pointnet~\cite{qi2017frustum} extracts the 3D frustum point clouds by extruding 2D bounding boxes from 2D image detectors. Then Pointnets~\cite{qi2017pointnet,qi2017pointnet++} are performed for each 3D bounding box frustum to get 3D points of interests. Finally, an amodal 3D box estimation network is applied to yield the 3D detection results.

\section{SIFRNet For 3D Object Detection}
In this section, we introduce the architecture of Scale Invariant and Feature Reweighting Network (SIFRNet) shown in Figure~1, which mainly composed of three parts: 3D instance segmentation network (Point-UNet), T-Net and 3D box estimation network (Point-SENet). Next, the design principles of SIFRNet architecture are introduced.

\begin{figure}[t]
\begin{center}
  \includegraphics[width=0.95\linewidth]{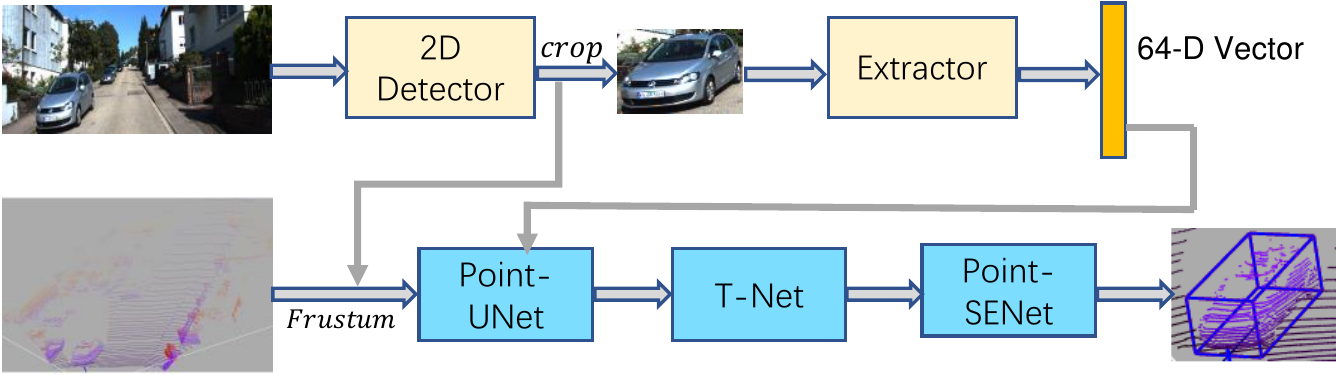}
\end{center}
\caption{The pipeline of SIFRNet for 3D object detection}
\label{fig:The pipeline of SIFRNet for 3D object detection}
\end{figure}

\begin{figure*}[htp]
\begin{center}
  \includegraphics[width=0.95\linewidth]{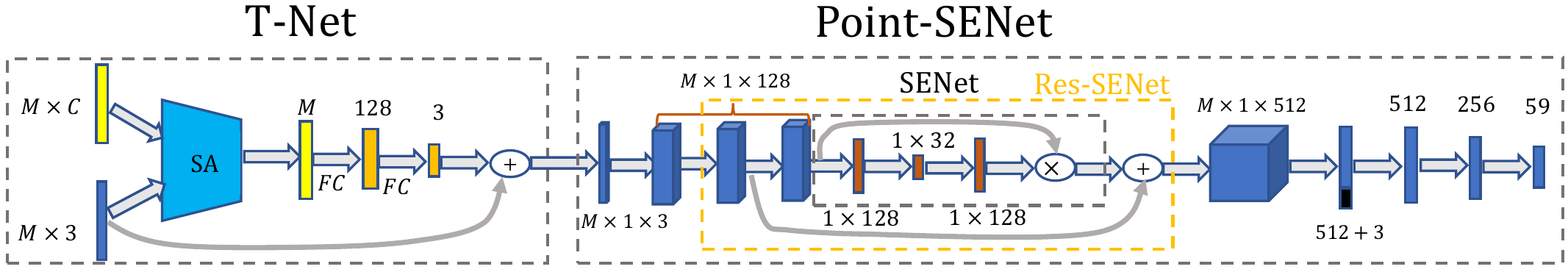}
\end{center}
\caption{The network architecture of T-Net and Point-SENet.}
\end{figure*}

\begin{figure}[t]
\begin{center}
  \includegraphics[width=0.95\linewidth]{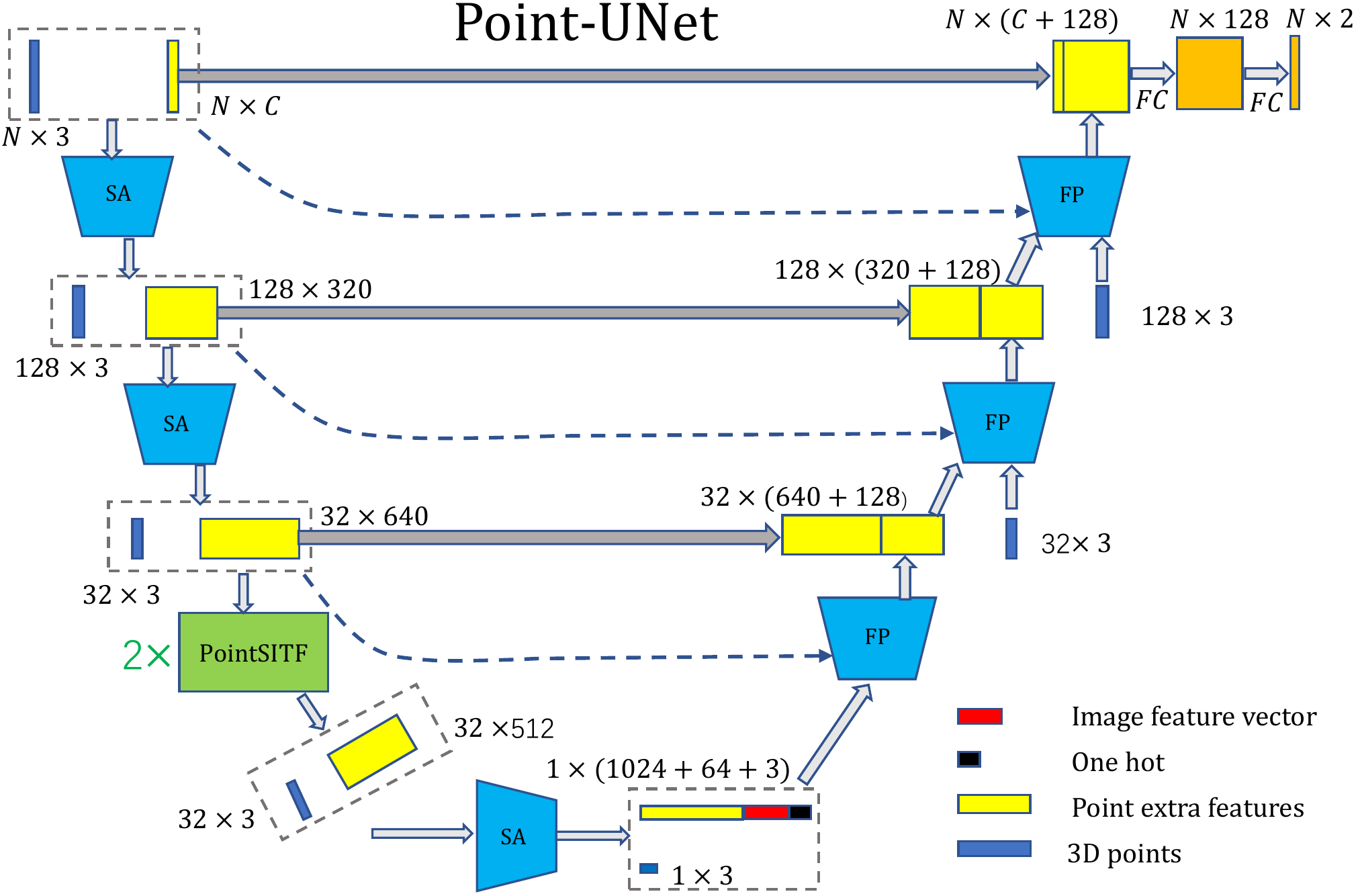}
\caption{The network architecture of Point-UNet.}
\end{center}
\end{figure}

\subsection{Point-UNet: 3D instance segmentation network}
This module takes 3D points in the frustum as the input and outputs the probability of each point to determine whether it is the point of interests or not. In the later 3D box estimation network, only these points of interests are really useful to the estimation of 3D bounding boxes, and other points may be the background or noise.

For one-class 3D detection tasks, an efficient model can be more easily designed to get satisfied detection results. However, objects in multiple categories have huge differences in size and orientation angle. These problems bring great difficulties for the 3D box estimation. Therefore, it is necessary that the network should have the ability to identify different types of objects. Then, the geometric features of objects in different categories can be distinguished when estimating 3D bounding boxes.

RGB image information with an effective and available feature plays an irreplaceable role in most 2D classification, detection and scene understanding tasks. For example, the color information is quite different from Cars, Pedestrians and Cyclists on the KITTI dataset. If the color information of these objects is known in advance, we can exploit the difference in color to get better performance. However, some objects are hardly to be distinguished only be the color information in sparse 3D point cloud space, such as Pedestrians and Cyclists. Luckily, the front view 2D image can provide much more appearance information, so we can encode the regions of interests into high-dimensional representations and fuse them into the network. Finally, the one-hot class information from 2D detection results can be utilized to improve the performance of the segmentation network as well.

The architecture of Point-UNet is shown in Figure~3. The input data of the network is $N\times3$ 3D point clouds for 3 axis, and the corresponding extra data is $N\times3$ RGB images for 3 channels and $N\times1$ Lidar reflection intensity maps. $N$ is the number of point clouds. In the beginning, the point cloud features can be extracted via two Set Abstract (SA) modules, which can be vividly understood as the process of Convolution and Pooling operations. Since Pointnets can not consider the orientation information of point clouds and the adaptability to the shape scales, we introduce the PointSIFT module based on the segmentation network of Pointnet++ into the network. Before the global SA module operation, two PointSIFT modules are integrated into our network. As a result, it has an ability to capture the orientation information in 3D space and has the robustness to different scale shapes. Then, we perform three FP operations, which can be considered as the process of Deconvolution and Upsampling operations. In addition, the previous point features at each stage of Feature Propagation (FP) layer are concatenated to gain richer information. Finally, the probability of the points of interests can be predicted via two fully-connected layers. The 3D instance segmentation network looks like an English letter U, so we name the network as Point-UNet.

\subsection{T-Net: 3D point cloud translation network}
The architecture of T-Net is shown in Figure~2, which is primarily used to estimate the center of a 3D box and translate the points of interests to the center of the box. After this process, the subsequent network only needs to further predict the residual for the final 3D box.

The input of the network is the points of interests from Point-UNet. The extracted feature is the reflection intensity of the Lidar. The input is imported into a SA module to extract the global features of 3D point clouds. Next, two fully-connected layers are used to predict the negative center coordinates, which are added to the original input points for achieving the process of the point cloud translation. The network is functionally similar to the translation network with similar residual network structure.

\subsection{Point-SENet: 3D box estimation network}
Point-SENet architecture is shown in Figure~2 which is mainly used to estimate the {$c_x$, $c_y$, $c_z$, $h$, $w$, $l$, $\theta$} parameters of a 3D box. $c_x$, $c_y$, $c_z$ represent the 3D box center coordinates along the X, Y, Z axis, respectively. $h$, $w$, $l$ represent the length, width and height of a 3D box, respectively. $\theta$ represents the heading angle along up-axis. In addition, the output of the network is a (3$+$4$\times$NS$+$2$\times$NH)-dimensional vector. NS denotes the number of size templates and NH represents for the number of orientation angle templates.

The input of the network is 3D point clouds from T-Net, no extra feature is used. At first, two point-wise convolutions are performed to get a high-dimensional point cloud feature which is $\mathbf{X_*}$. Since the general convolutional neural networks do not consider the relationship of all channel features, it may causes some useless features propagating to subsequent networks, decreasing the performance of the entire network. In order to improve the feature representation ability of the network, the SENet module is extended into 3D space for operating on 3D point clouds directly.

The third convolutional operation in Point-SENet model is used to obtain the input feature of SENet module $\mathbf{X}$ . First, the squeeze operation is performed by a global max pooling layer instead of the global average pooling layer to aggregate spatial information into the channel feature which is formulated as follows:
\begin{equation}
\mathbf{C}_{sq}^l = \max (\mathbf{X}_{1:h,1:w}^l)
\end{equation}
where $h=M$ represents the number of 3D points of interests. $w$ is equal to $1$ and $l={1,...,c}$ where $c$ is the number of channel features in our model. Second, the excitation operation is performed via  two fully-connected layers and one sigmoid function. Through the above operations, we can obtain the correlation between all channel features. The excitation operation is used to obtain the channel-wise dependency and formulated as:
\begin{equation}
\mathbf{Scale}^l = Sigmoid(\mathbf{W}_2Relu(\mathbf{W}_1\mathbf{C}_{sq}^l))
\end{equation}
where $\mathbf{W}_1 \in \mathcal{R}^{\frac{c}{r}\times{c}}$  and $\mathbf{W}_1 \in \mathcal{R}^{c \times \frac{c}{r}}$ represent the parameters of two fully connected layers, respectively. $r$ denotes the reduction rate of the bottleneck ($r=4$ in our model). Next, the original feature is multiplied by $\mathbf{Scale}^l$, obtaining the feature $\mathbf{X}_{se}^l$, which achieves the process of reweighting the feature channels of 3D point clouds.
\begin{equation}
\mathbf{X}_{se}^l = \mathbf{Scale}^l \cdot \mathbf{C}_{ex}^l
\end{equation}
Since the value of $\mathbf{Scale}^l$ is limited to $(0,1)$, the gradient disappears easily during the back-propagation, which makes the optimization of the network very difficult. Encouraged by the residual networks~\cite{he2016deep}, we use the following formulation for better gradient back-propagation:
\begin{equation}
\mathbf{F}_{SE}^l = \mathbf{X}_{se}^l + \mathbf{X}_*^l
\end{equation}
The model can automatically obtain the importance of channel features by self-learning. According to the importance of each channel feature, useful features are enhanced while features that have little information to the network are suppressed. The final output feature via SENet module can be represented as $\mathbf{F}_{SE} = \{\mathbf{F}_{SE}^1,\mathbf{F}_{SE}^2,...,\mathbf{F}_{SE}^c\}$.
After that, a point-wise convolution operation is performed to upgrade the reweighted features to 512 dimensions. It is aimed at alleviating the loss of information in the subsequent pooling operation. Finally, three fully-connected layers are used to estimate all residual parameters of 3D bounding boxes.

\begin{table*}[h]
\scriptsize
\centering
\begin{tabular}{l|c|c|c|c|c|c|c|c|c|c}
\hline
\multirow{2}{*}{Method} & \multicolumn{3}{c|}{Cars} &
\multicolumn{3}{c|}{Pedestrians} & \multicolumn{3}{c|}{Cyclists} & \multicolumn{1}{c}{\multirow{2}{*}{3D mAP}} \\
\cline{2-10}

& Easy & Moderate & Hard & Easy & Moderate & Hard & Easy & Moderate & Hard & \multicolumn{1}{c}{} \\
\hline
\hline
Mono3D~\cite{chen2016monocular} &2.53 &2.31 &2.31 &- &- &- &- &- &- &- \\
3DOP~\cite{chen20153d}   &6.55 &5.07 &4.10 &- &- &- &- &- &- &- \\
VeloFCN~\cite{li20173d}	 		            & 15.20 & 13.66 & 15.98 & - & - & - & -  & - & - & - \\
MV3D~\cite{chen2017multi} 	                    & 71.29 & 62.68 & 56.56 & - & - & - & -  & - & - & - \\
Pointfusion-final~\cite{xu2017pointfusion}  			& 77.92 & 63.00 & 53.27 & 33.36 & 28.04 & 23.38 & 49.34  & 29.42 & 26.98 & 42.75\\
AVOD(Feature Pyramid)~\cite{ku2018joint}   &84.41  &\textbf{74.44}  &\textbf{68.65}   &-  &58.8  &-  &- &49.7 &-  &-\\
F-pointnet(V1)~\cite{qi2017frustum}				    & 83.26 & 69.28 & 62.56 & - & - & - & -  & - & - & -\\
F-pointnet(V2)~\cite{qi2017frustum}	 			    & 83.76 & 70.92 & 63.65 & \textbf{70.00} & \textbf{61.32} & \textbf{53.59} & 77.15  & 56.49 & 55.37 & 65.58\\
\hline
V2-SENet			        & 84.71 & 71.25 & 63.74 & 69.14 & 60.12 & 52.91 & 78.71  & 57.43 & 53.55 & 65.73\\
V2-SENet-PointSIFT			& 84.45 & 71.87 & 64.06 & 69.13 & 60.21 & 53.10 & 79.43  & 58.55 & 54.86 & 66.18\\
V2-SENet-PointSIFT-rgb-image   &\textbf{85.99} &72.72 &64.58 &69.26 &60.54 &52.90 &79.43 &59.26 &55.09 &66.64 \\
Fine-tune-final 	& 85.62 &72.05 &64.19 & 69.35 & 60.85
                & 52.95 & \textbf{80.87}  & \textbf{60.34} & \textbf{56.69} & \textbf{66.99}\\
\hline
\end{tabular}
\caption{$AP_{3D}$ (\%) results on KITTI validation set for 3D object detection.}
\label{T1}
\end{table*}

\begin{table*}[h]
\scriptsize
\centering
\begin{tabular}{l|c|c|c|c|c|c|c|c|c|c}
\hline
\multirow{2}{*}{Method} & \multicolumn{3}{c|}{Cars} &
\multicolumn{3}{c|}{Pedestrians} & \multicolumn{3}{c|}{Cyclists} & \multicolumn{1}{c}{\multirow{2}{*}{3D mAP}} \\
\cline{2-10}

& Easy & Moderate & Hard & Easy & Moderate & Hard & Easy & Moderate & Hard & \multicolumn{1}{c}{} \\
\hline
\hline
Mono3D~\cite{chen2016monocular} &5.22 &5.19 &4.13 &- &- &- &- &- &- &-  \\
3DOP~\cite{chen20153d}   &12.63 &9.49 &7.59 &- &- &- &- &- &- &-  \\
VeloFCN~\cite{li20173d}		            & 40.14 & 32.08 & 30.47 & - & - & - & -  & - & - & - \\
MV3D~\cite{chen2017multi} 	                    & 86.55 & 78.10 & \textbf{76.67} & - & - & - & -  & - & - & - \\
Pointfusion-final~\cite{xu2017pointfusion} 			& 87.45 & 76.13 & 65.32 & 37.91 & 32.35 & 27.35 & 54.02  & 32.77 & 30.19 & 49.28\\
F-pointnet(V1)~\cite{qi2017frustum}			    & 87.82 & 82.44 & 74.77 & - & - & - & -  & - & - & -\\
F-pointnet(V2)~\cite{qi2017frustum} 			    & 88.16 & \textbf{84.02} & 76.44 & 72.38 & 66.39 & 59.57 & 81.82  & 60.03 & 56.32 & 71.68\\
\hline
V2-SENet			        & 88.43 & 82.61 & 75.71 & 72.66 & 66.75 & 59.78 & 83.36  & 63.39 & 58.98 & 72.41\\
V2-SENet-PointSIFT			& 88.19 & 83.62 & 76.23 & 72.61 & 66.82 & 59.38 & 83.54  & 64.13 & 59.16 & 72.63\\
V2-SENet-PointSIFT-rgb-image &\textbf{88.80} &83.96 &76.21 &75.39 &67.51 &60.03 &83.01 &64.02 &59.51 &73.16 \\
Fine-tune-final	& 88.63 & 83.45 & 76.08 & \textbf{76.67} & \textbf{68.49}
                & \textbf{60.78} & \textbf{83.50}  & \textbf{64.57} & \textbf{59.87} & \textbf{73.56}\\
\hline

\end{tabular}
\caption{$AP_{2D-Bird-View}$ (\%) results on KITTI validation set for 3D object localization.}
\label{T2}
\end{table*}

\subsection{Loss Function}
We compute the loss function for Point-UNet, T-Net and Point-SENet jointly.
$L_{seg}$ is the classification loss for Point-UNet. $L_{T-Net}$ is
the center regression loss for T-Net. For Point-SENet, $L_{c-reg}$ and $L_{angle-cls}$ are losses for estimation of the orientation angle. $L_{angle-reg}$ and $L_{size-clc}$ are losses for the estimation of 3D box size. $L_{corner}$ represents the distances between the eight corners of the groundtruth 3D box and the predicted 3D box. The corner loss can refer to~\cite{qi2017frustum} for details. The total loss function can be formulated as follows:
\begin{equation}
\begin{aligned}
 L_{total} & =  L_{seg} + L_{T-Net} + \lambda (L_{center-reg}+ \\
 & L_{angle-cls}+ L_{angle-reg}+L_{size-clc} \\
  & +L_{size-reg} + \gamma L_{corner})
\end{aligned}
\end{equation}

In addition, the new L2 loss function is redefined for $L_{angle-reg}$. The L2 loss for an orientation angle is equal to the Euclidean distance between the true angle and the predicted angle, which is a more reliable angle regression for estimating 3D bounding boxes. It is worth noting that cosine function is not easy to optimize, so we only use it in the process of fine-tuning.
\begin{equation}
\begin{aligned}
L_{angle-reg} & = \frac{1}{B} \sum_{m=1}^{NS} \sum_{n=1}^{NH} {M}_{mn} |e^{j                  \theta_{mn}}-e^{j\theta_{mn}^*}|^2  \\
              & = \frac{1}{B} \sum_{m=1}^{NS} \sum_{n=1}^{NH} {M}_{mn}
              [(cos\theta_{mn}-cos{\theta_{mn}^*})^2 \\
              &+ (sin\theta_{mn}-sin{\theta_{mn}^*})^2 ]\\
              & = \frac{1}{B} \sum_{m=1}^{NS} \sum_{n=1}^{NH} {M}_{mn}[2-2cos(\theta_{mn}-\theta_{mn}^*)]
\end{aligned}
\end{equation}
 where $B$ is the batch size and $M$ is the mask which can be obtained from the output of Point-UNet. $\theta$ represents the groundtruth orientation angle and ${\theta^*}$ means the predicted orientation angle.

\section{Experiments}
\subsection{Datesets and Evaluation Metrics}
We conduct our experiments on KITTI and SUN-RGBD datasets. 3D IoU (Intersection over Union) is used as the common evaluation metric. If 3D IoU with the groundtruth is over a given threshold, the predicted 3D box is considered as a true positive (TP). For KITTI dataset, we follow the official KITTI evaluation protocol, where 3D IoU threshold is 0.7 for Cars category and 0.5 for both Pedestrians and Cyclists categories, respectively. On SUN-RGBD dataset, the 3D IoU is 0.25 for all the categories.

{\bf KITTI:} The KITTI 3D object detection dataset contains 7481 training images and Velodyne Lidars. We follow the settings in ~\cite{qi2017frustum} and split the dataset into training set and validation set with 3717 and 3769 samples, respectively. Finally, we report the results on the validation set for Cars, Pedestrians and Cyclists categories.

{\bf SUN-RGBD:} The SUN-RGBD dataset have 700 object categories, the training set and test set contains 5285 and 5050 images, respectively. But we only report the results of ten kinds of objects in the test set with the same settings in~\cite{qi2017frustum}. This is because there are more samples on these categories and the object sizes are also relatively larger than the others.

\begin{figure*}[t]
\begin{center}
  \includegraphics[width=0.87\linewidth]{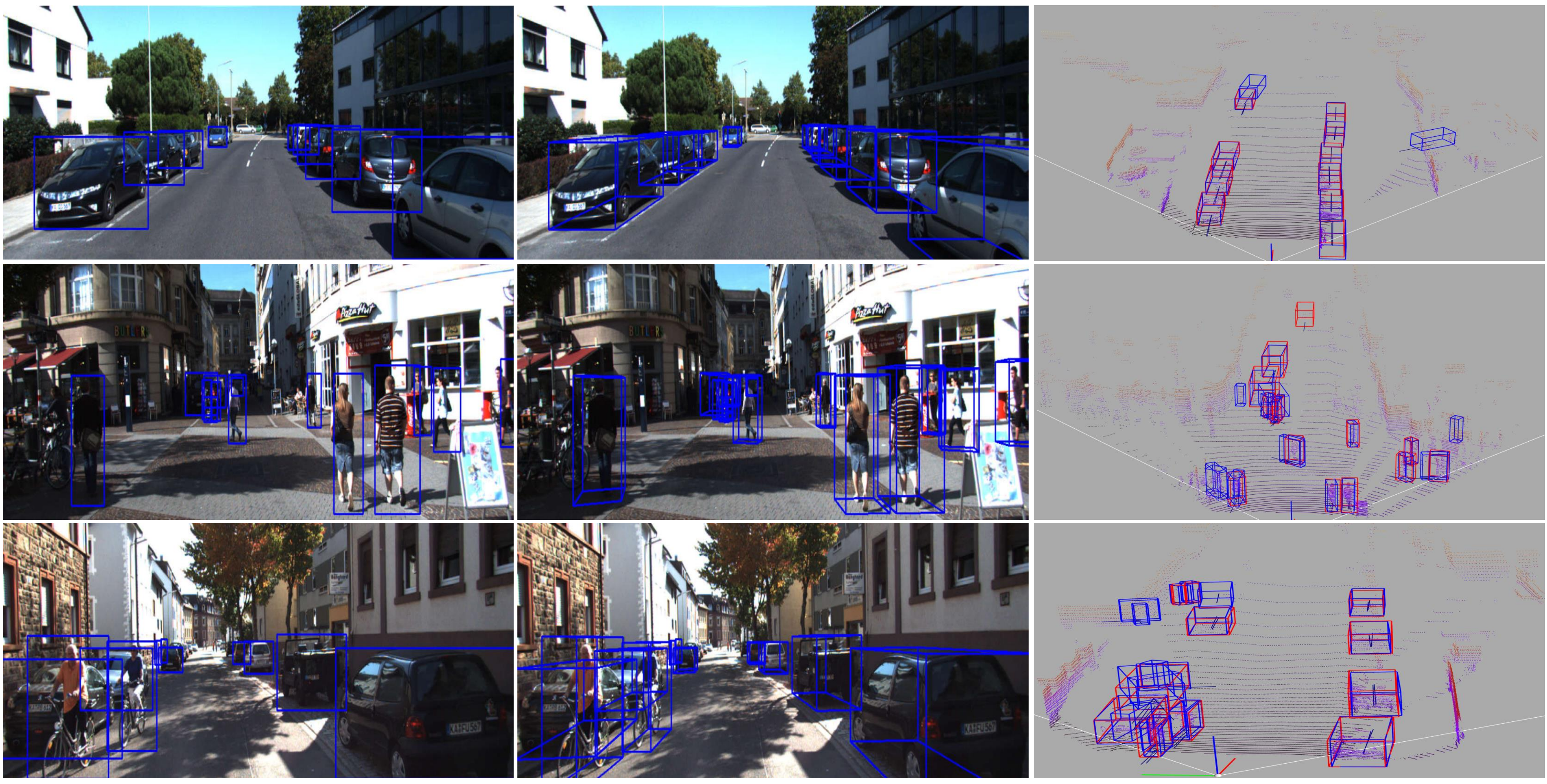}
\end{center}
\caption{The visualization results for 3D object detection on KITTI validation set. The first column shows the 2D detection results on 2D images. The second column displays the 3D boxes that are projected on the 2D image. The last column demonstrates our 3D detection results on Lidar. The groundtruths are in red and the predicted boxes are in blue.}
\end{figure*}

\begin{figure}[h]
\begin{center}
  \includegraphics[width=0.95\linewidth]{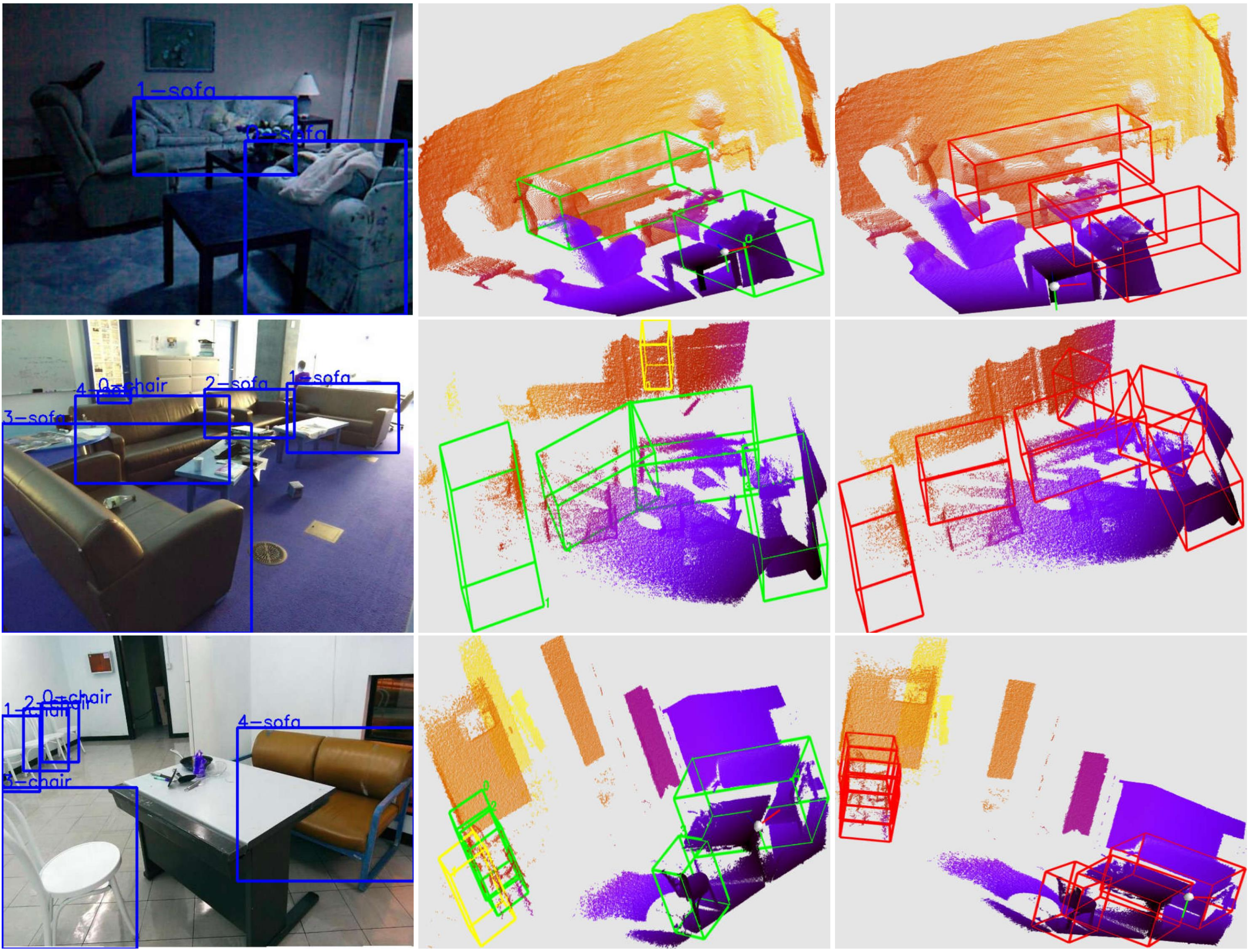}
\end{center}
\caption{The visualization results for 3D object detection on SUN-RGBD test set. The first column shows the 2D bounding boxes based on YoloV3. The second column displays the 3D bounding boxes on 3D point clouds. The green boxes are true positives and yellow boxes are false positives. The 3D IoU threshold is 0.25. The last column illustrates the groundtruths which are in red.}
\end{figure}

\subsection{Implementation Details}
{\bf The details on 2D detector:} In 2D detection tasks, manually selecting anchor boxes can not match the distribution of the 2D boxes well from the training set, and it is better to apply K-means~\cite{jain2010data} to cluster the anchor boxes from the training set. YoloV3~\cite{redmon2018yolov3} is used as the 2D detector to generate the final 2D bounding boxes. The model is pre-trained on MS-COCO dataset and fine-tuned on SUN-RGBD dataset. We adopt Adaptive Moment Estimation (Adam) to optimize the model. The learning rate is equal to 1e-4. During the training process, we freeze the first 185 layers and release all the layers after 50 epochs. The training is terminated after 100 epochs. For fair comparisons, 2D detection results are the same with~\cite{qi2017frustum} on KITTI dataset.

{\bf The details on SIFRNet:} The input data sizes of Point-UNet, T-Net and Point-SENet are $N{\times}7$, $N{\times}4$ and $N{\times}3$, respectively. ResNet-50 is used to extract the features of the cropped images from 2D detector. The 64-dimensional feature vector is obtained by the AveragePooling2D layer of ResNet-50. The Adam optimizer is adopted to optimize the deep neural networks with the learning rate of 0.001. The first exponential decay rate is equal to 0.95 and the second exponential decay rate is equal to 0.999. In addition, data augmentation on point clouds is performed in the following two ways: randomly flip the point clouds in the frustums and shift the point clouds in the frustums along the z-axis direction.

\subsection{Comparisons to the state-of-the-art methods}
{\bf Experimental results on KITTI:} The results of our 3D object detection on KITTI validation set are shown in Table~1. Comparing with MV3D~\cite{chen2017multi} and Pointfusion~\cite{xu2017pointfusion} that use multi-sensor fusion strategy, our network greatly outperforms these state-of-the-art methods. In the experiments, our V2-SENet method only uses SENet module instead of the three SA modules on F-Pointnet (v2)~\cite{qi2017frustum} which contains more parameters. With the reduced number of parameters of our model, the performance still has a slight improvement. V2-SENet-PointSIFT is based on the V2-SENet model with an additional PointSIFT module. V2-SENet-PointSIFT-rgb-image is the model that image features and RGB information are attached to V2-SENet-PointSIFT for further improvements. Fine-tune-final is the model that we fine-tune on the V2-SENet-PointSIFT-rgb-image model by using the redefined angle loss function. 
As shown in the results, our methods have a lower accuracy on the Cars class than AVOD for 3D object detection. It should be noted that our network generates a single model to predict all categories for a good generalization ability. However, AVOD separates Cars category individually for training to achieve better performance. This makes the trained model not adaptive for all object categories.

In order to illustrate the effectiveness of the new angle regression loss, we use the same pre-trained model to experiment on different angle loss functions with various epochs. The 3D mAP results of the original loss are 65.37\%, 65.82\%, 66.54\% and 66.68\% (epochs=1, 5, 10 and our fine-tune-final result) while the corresponding 3D mAP results of the proposed loss are 65.61\%, 66.25\%, 66.91\% and 66.99\%, respectively. Based on the above results, our angle loss has more contributions on the performance. 
In addition, Table~2 also reports the results on 3D object localization. Since Cyclists category contains persons, it is very difficult to distinguish Cyclist and Pedestrian categories. Nevertheless, our method still outperforms the cutting-edge methods on Cars and Cyclists categories. Figure~4 shows the visualization results of our method for 3D object detection on KITTI validation set.

{\bf Experimental results on SUN-RGBD:} Different from the KITTI dataset for outdoor scenes with only three types of objects, SUN-RGBD contains much more kinds of objects in indoor environment which bring us even greater challenges for 3D object detection. In the experiments, YoloV3 is used as our 2D detector and has achieved 2D mAP of 53.9\% on SUN-RGBD test dataset. The whole network architecture is the same to the one on KITTI dataset. Table~3 indicates that our method significantly outperforms the state-of-the-art methods which include DSS~\cite{song2016deep}, 2d-driven~\cite{lahoud20172d}, Pointfusion~\cite{xu2017pointfusion}, COG~\cite{ren2016three} and F-Pointnet (v1)~\cite{qi2017frustum}. Surprisingly, We achieve 4.4\% 3D mAP higher than F-Pointnet~\cite{qi2017frustum}. For 2048 input 3D points, our method has a relative 2.3\% 3D mAP higher than V1-2048. Figure~5 shows the visualization results of our method for 3D object detection on SUN-RGBD test set.

\begin{table*}[t]
\scriptsize
\centering
\begin{tabular}{l|c|c|c|c|c|c|c|c|c|c|c}
\hline
 Method & bathtub  & bed  & bookshelf & chair & desk & dresser & nightstand & sofa  &  table  & toilet  &3D mAP \\
\hline
\hline
    DSS~\cite{song2016deep}   & 44.2 & 78.8 & 11.9 & 61.2 & 20.5 & 6.4 & 15.4 & 53.5 & 50.3 & 78.9 & 42.1\\
    2d-driven~\cite{lahoud20172d}  & 43.5 & 64.5 & 31.4 & 48.3 & 27.9 & 25.9 & 41.9 & 50.4 & 37.0 & 80.4 & 45.1\\
    Pointfusion~\cite{xu2017pointfusion}  & 37.3 & 68.6 & 37.7 & 55.1 & 17.2 & 24.0 & 32.3 & 53.8 & 31.0 & 83.8 & 44.1 \\
    COG~\shortcite{ren2016three}  & 58.3  & 63.7 & 31.8 & 62.2 & 45.2 & 15.5 & 27.4 & 51.0 & 51.3 & 70.1 & 47.6 \\
    F-Pointnet(v1)~\cite{qi2017frustum}  & 43.3  & 81.1 & 33.3 & 64.2 & 24.7 & 32.0 & 58.1 & 61.1 & 51.1 & 90.9 & 54.0 \\
    \hline
    V1-1024    & 51.6 & 82.0 &32.2 & 54.6 &33.5 &32.4 & 67.9 &66.3 & 48.0 &\textbf{88.2} &55.7\\
    V1-2048   & 51.6 & 83.1 &35.4 & 54.5 &33.0 &\textbf{33.7} &\textbf{68.2} & 66.7 &48.2 &87.1 &56.1\\
    Ours-1024   & 61.5 & 83.3 & 38.1 & 57.7 &33.8 &32.8 &67.3 &\textbf{67.4} &51.3 &87.3 &58.1\\
    Ours-2048   &\textbf{64.0}  &\textbf{84.4} & \textbf{38.4} &\textbf{57.9} & \textbf{34.1} &32.2 &67.7 &67.3 &\textbf{51.4} &86.2 &\textbf{58.4} \\
\hline
\end{tabular}
\caption{Comparison with the state-of-the-art methods on SUN-RGBD test dataset.}
\label{T3}
\end{table*}

\subsection{Influence of 2D Detection}

In order to analyze the influence of 2D detection on the final 3D detection, we choose several trained models of YoloV3 to get different 2D detection performance on SUN-RGBD test dataset. In Table~4, it shows that our method can get 2.0\% to 2.6\% 3D mAP higher than the original F-Pointnet (V1) with the same 2D detection results, respectively. The results demonstrate that the higher 2D performance comes the bigger gain of our model for 3D detection.

\begin{table}[h]
\scriptsize
\begin{center}
\begin{tabular} {l|c|c|c}
\hline
\multirow{2}{*}{2D mAP} & \multicolumn{3}{c}{3D mAP} \\
\cline{2-4}& V1 & Our  &Gain\\
\hline
\hline
40.6 		     & 46.1 &\textbf{48.1}   &2.0  \\
47.1 		     & 49.6 &\textbf{51.6}   &2.0  \\
50.5 	         & 52.2 &\textbf{54.3}   &2.1  \\
52.9 	         & 55.5 &\textbf{57.6}   &2.1  \\
53.9 			 & 56.1 &\textbf{58.4}   &2.3  \\
GT			     & 84.1 &\textbf{86.7}   &2.6  \\
\hline
\end{tabular}
\end{center}
\caption{The influence of 2D detection.}
\label{tab:Influence of 2D Detection}
\end{table}

\subsection{Influence of the Number of Input Points}
The KITTI dataset is collected from a long-distance outdoor scene, sometimes there are only a few points on the 3D object. To analyze the influence of the number of 3D points for 3D bounding box estimation, we perform the experiments on SUN-RGBD test dataset which has reliable dense depth maps. We take 32, 128, 256, 512, 1024, 2048 points (in a frustum) as the input, respectively. Table~5 shows 3D mAP results with different number of input points. It can be seen that our model can achieve a big improvement when the number of input points is very small. In particular, with 32 3D points as the input, our mAP still obtains 51.1\%,  which is 7.2\% higher than the V1 result with a huge margin. It can be proved that our model is quite suitable for highly sparse 3D point clouds.

\begin{table}[h]
\scriptsize
\begin{center}
\begin{tabular} {l|c|c|c|c|c|c}
\hline
\multirow{2}{*}{\shortstack{Number of\\input points}} & \multicolumn{6}{c}{3D mAP} \\
\cline{2-7}& V1$_{53.9}$ & Our$_{53.9}$ & Gain & V1$_{GT}$ & Our$_{GT}$ &Gain\\
\hline
\hline
32 		   &43.9 &\textbf{51.1} &7.2 &63.1 &\textbf{70.2} &7.1      \\
128 	   &53.1 &\textbf{56.3} &3.2 &79.2 &\textbf{83.2} &3.0      \\
256 	   &54.4 &\textbf{56.9} &2.5 &81.4 &\textbf{85.3} &2.9	   \\
512		   &55.4 &\textbf{57.8} &2.4 &83.2 &\textbf{86.0} &2.8     \\
1024 	   &55.7 &\textbf{58.1} &2.4 &83.7 &\textbf{86.3} &2.6	   \\
2048       &56.1 &\textbf{58.4} &2.3 &84.1 &\textbf{86.7} &2.6     \\
\hline
\end{tabular}
\end{center}
\caption{The influence of the number of input points.}
\label{tab:Influence of the Number of Input Points}
\end{table}

\section{Conclusion}
In this paper, SIFRNet is put forward for 3D object detection, which is suitable for both indoor and outdoor scenes. The proposed architecture can make full use of the advantages of both RGB images and 3D point clouds. Outstanding experiment results both on KITTI dataset which contains a large number of sparse point clouds and SUN-RGBD dataset which contains many occluded objects reveal that our model has a certain generalization ability and robustness on 3D object detection tasks. Even when the point clouds are extremely sparse, our method can still obtain very satisfied results, which also demonstrates that our model can provide a better 3D representation. In future work, we will focus on the end-to-end trainable model for Lidar-only 3D data to improve the efficiency of 3D object detection tasks.

\section{Acknowledgement}
This project was partially supported by the National Natural Science Foundation of China (Grant No.61602485 and Grant No.61673375), the National Key Research and Development Program of China (Grant No.2016YFB1001005), and the Projects of Chinese Academy of Sciences (Grant No.QYZDB-SSW-JSC006 and Grant No.173211KYSB20160008).

\bibliography{camera-ready}
\bibliographystyle{aaai}
\end{document}